\documentclass[journal]{IEEEtran}
\usepackage{diagbox}
\usepackage{multirow}
\usepackage{times}
\usepackage[table]{xcolor}
\usepackage{booktabs,array,xcolor,colortbl,multicol,times,epsfig,graphicx,subfigure}
\usepackage{amsmath,amsthm,amsfonts,amssymb,bm}
\usepackage[vlined,boxed,commentsnumbered,ruled,linesnumbered]{algorithm2e}
\usepackage{caption}
\usepackage{colortbl,accents}
\usepackage{subfigure}
\usepackage{caption}
\def\widebar{\accentset{{\cc@style\underline{\mskip10mu}}}}
\def\Widebar{\accentset{{\cc@style\underline{\mskip8mu}}}}
\makeatother

\title{Decoupled Learning for Factorial Marked Temporal Point Processes}
\begin{document}

\author{Weichang~Wu,
	Junchi~Yan*,~\IEEEmembership{Member,~IEEE,}
	Xiaokang~Yang,~\IEEEmembership{Senior~Member,~IEEE,}
	Hongyuan~Zha
	\thanks{W. Wu and X. Yang are with the Department of Electrical Engineering, Shanghai Jiao Tong University, Shanghai, 200240, China. E-mail: blade091@sjtu.edu.cn,	xkyang@sjtu.edu.cn.}
	\thanks{J. Yan (correspondence author) is with the Department of Computer Science and Engineering, Shanghai Jiao Tong University, Shanghai, 200240, China. E-mail: yanesta13@163.com.}
	\thanks{H. Zha is with School of Computational Science and Engineering, College of Computing, Georgia Institute of Technology, Atlanta, Georgia, 30332, USA and East China Normal University, Shanghai, 200062, China. E-mail: zha@cc.gatech.edu}}

\maketitle
\begin{abstract}
This paper introduces the factorial marked temporal point process model and presents efficient learning methods. In conventional (multi-dimensional) marked temporal point process models, event is often encoded by a single discrete variable i.e. a marker. In this paper, we describe the factorial marked point processes whereby time-stamped event is factored into multiple markers. Accordingly the size of the infectivity matrix modeling the effect between pairwise markers is in power order w.r.t. the number of the discrete marker space. We propose a decoupled learning method with two learning procedures: i) directly solving the model based on two techniques: Alternating Direction Method of Multipliers and Fast Iterative Shrinkage-Thresholding Algorithm; ii) involving a reformulation that transforms the original problem into a Logistic Regression model for more efficient learning. Moreover, a sparse group regularizer is added to identify the key profile features and event labels. Empirical results on real world datasets demonstrate the efficiency of our decoupled and reformulated method. The source code is available online.
\end{abstract}


\section{Introduction and Background}
Events are ubiquitous across different domains and applications. In e-commerce, events refer to the transactions associated with users, items. In health informatics, event sequence can be a series of treatment over time of a patient. In predictive maintenance, events can carry important log data for when the failure occurs and what is the type. In all these examples, effectively modeling and predicting the dynamic behavior is of vital importance for practical usefulness.

\textbf{Marked temporal point process}
Point process \cite{daley2007introduction} is a useful tool for modeling the event sequence with arbitrary timestamp associated with each event. An event in point process can carry extra information called marker. The marker typically refers to event type and lies in the discrete label space i.e. a finite category set $\{1, ..., m\}$\footnote{A general concept can be found in \cite{Gelfand2010SS}: \emph{a marked point pattern is one in which each point of the process carries extra information called a mark, which may be a random variable, several random variables, a geometrical shape, or some other information}. In this paper, we focus on discrete labels for marks. The marked point process is also termed by \emph{multi-dimensional point process} \cite{LinigerPhD2009}, where each dimension refers to a discrete mark value.}.

\textbf{Factorial marked temporal point process}
For the above mentioned marked point process, the event is represented by a single mark as a single discrete variable, but in many application scenarios, the event can carry multiple markers. For instance, a movement to a new job carries both the label for position and label for company, which can be treated by two orthogonal markers with different values. Though such cases are ubiquitous in real world, the factorial marked point processes have drawn little attention in literature as existing literatures mostly work with a single marker \cite{LinigerPhD2009,ZhouAISTATS13,XiaoAAAI17}. Inspired by Factorial Hidden Markov Models \cite{ghahramani1996factorial}, we introduce the factorial marked temporal point process, in which the event is represented by multiple markers, and propose a decoupling method to learn the process.



\textbf{Intensity function and problem statement}
One core concept for point process is the intensity function $\lambda(t)$, which represents the expected instantaneous rate of events at time $t$ conditioned on the history. One basic intensity function is the constant $\lambda(t)=\lambda_{0}$ over time, as used in the homogeneous Poisson process. Another popular form is the one used by the Hawkes process \cite{HawkesBiometrika71}: $\lambda(t)=\gamma_0+\alpha\sum_{t\in \tau}\gamma(t,t_i)$, where $\tau$ denotes the event history and $\gamma(t,t_i)\geq 0$ is a marker-vs.-marker infectivity kernel capturing the temporal dependency between event at $t$ and at $t_i$. 

In this paper, we are interested in describing a factorial marked point process for event marker prediction task by using the history event information and the individual level profile of an event taker. We focus on the next-event label estimation, distributed over more than one markers. In particular, our empirical study focuses on individual level next job prediction involving both position and company for LinkedIn users, and duration prediction in current ICU department and transition prediction to next ICU department for patients in MIMIC-II database \cite{goldberger2000physiobank}.


\section{Related Work and Contribution}
\textbf{Learning for temporal point process}
Point process is a powerful tool for modeling event sequence with timestamp in continuous time space. Early work dates back to the Hawkes processes \cite{HawkesBiometrika71} which shows appropriateness for self-exciting and mutual-exciting process like earthquake and its aftershock \cite{OgataJASA88,OgataJASA98}. The learning is fulfilled by maximum likelihood estimation by directly computing the gradient and Hessian matrix w.r.t. the log-likelihood function \cite{ozaki1979maximum}. Recently more modernized machine learning approaches devise new efficient algorithms for learning the parameters of the specified point process. Nonparametric Expectation-Maximization (EM) algorithm is proposed in \cite{LewisJNS2011} for multiscale Hawkes Processes using the majorization-minimization framework, which shows superior efficiency and robustness compared with sampling based estimation methods. \cite{ZhouAISTATS13} extends the technique to handle the multi-dimensional Hawkes process by adding a low-rank and sparsity regularization term in the maximum likelihood estimation (MLE) based loss function.

\textbf{Factorial model} Though almost all of these works mentioned above involve the infectivity matrix for model parameters learning, none of them considers the factorial temporal point process case, i.e. an event type is factored into multiple markers, which leads to the explosion of the infectivity matrix size. The idea of factorizing events or states into multiple variables is employed in \cite{ghahramani1996factorial} for Hidden Markov Models (HMM) using variational methods to solve data mining task like capturing statistical structure, but little literature is found about its utility in point process. To our best knowledge, this is the first work of factorial marked point process learning for event marker prediction. Note timestamp prediction can be approximated by predicting a predefined time interval as time duration marker as done in this paper.


\textbf{Sparse regularization for point process}
Sparse regularization is a well-established technique in traditional classification and regression models, such as the $\ell_1$ regularizer \cite{ng2004feature}, group Lasso \cite{meier2008group}, sparse group regularizer \cite{simon2013sparse}, etc. Recent point process models have also found their applications like the $\ell_1$ regularization used in \cite{li2014learning} to ensure the sparsity of social infectivity matrix, the nuclear norm in \cite{ZhouAISTATS13} to get a low-rank network structure and the group Lasso in \cite{xu2016icu} for feature selection. We propose to use the sparse group regularizer, which encourages the nonzero elements focusing on a few columns obeying with the intuition that only a few features and labels play the major role in event dynamics. We find little work in literature on group sparse regularizer for point process learning.

\textbf{Contributions} The main contributions of this paper are:

1) We introduce the concept of factorial marked point process for event marker prediction, and propose a decoupled learning algorithm to simplify the factorial model by decoupling the marker mutual relation in modeling. The method outperforms general marked point process on real-world datasets.


2) We present a multi-label Logistic Regression (LR) perspective and devise reformulation towards a class of point process discriminative learning problems. It eases the learning of these processes by using on-the-shelf LR solver.

Besides these major contributions, we also make additional improvements in proposing a regularized learning objective, which we will include for completeness.

\section{Proposed Model and Objective Function}
\subsection{Factorial point process}
Factorial point process refers to the processes in which event can be factorized into multiple markers. Except the \emph{job movement prediction} and \emph{ICU department prediction} mentioned in Introduction, many application cases can be described by the factorial point process, while haven't been explored yet. For instance, a \emph{weather forecast} containing \emph{temperature}, \emph{humidity}, \emph{precipitation}, and \emph{wind} can be seen as a factorial point process with $4$ markers, with each marker having discrete or continuous values. Obviously these factors affect each other, e.g. the \emph{humidity} today is influenced by the \emph{precipitation} and \emph{temperature} in recent few days. The conventional marked point process could only model one of these factors using a single marker without considering the infectivity between these factors. A factorial point process with multiple markers for the event is essential.

Learning factorial point process is challenging. Taking \emph{job movement prediction} with two markers \emph{company} $c$ and \emph{position} $p$ as example: to predict the probability of user $x$'s $n$-th job $(c_n,p_n)$, we need to learn a $4$-dimension tensor representing the impact of history \emph{companies} $\{c_i\}_{i=1}^{n-1}$ on $c_n$ and $p_n$, the impact of history \emph{positions} $\{p_i\}_{i=1}^{n-1}$ on $c_n$ and $p_n$, respectively. In point process, it means we need to learn a set of intensity functions including $\lambda(c,c)$, $\lambda(c,p)$, $\lambda(p,c)$ and $\lambda(p,p)$. This simple case considers no infectivity between different sequences, i.e., if we also consider the impact of another user \emph{y}'s job movement on user $x$'s choice of $c_n$ and $p_n$, we would compute a $6$-dimension tensor to measure the complete infectivity, with two extra intensities $\lambda(y,c)$ and $\lambda(y,p)$.

There are ways to simplify factorial point process learning, e.g. we can treat the combination of multiple factors as one marker, and use the conventional marked point process model, but this will lead to explosion of the size of infectivity matrix. In this paper, 
we explore a simple decoupling solution that decouple the factorial point process into separate models of different markers respectively. As shown in Fig.\ref{fig:matrix_a} for the instance of $2$ markers $c$ and $p$, we decouple the original infectivity matrix into smaller one by introducing $4$ tensor variable $\bm{a}_{pp}$, $\bm{a}_{pc}$, $\bm{a}_{cp}$ and $\bm{a}_{cc}$. We will present the decoupled model in details in the following section.

\subsection{Decoupled learning for factorial point process}
More generally, we discuss the situation that event can be factorized into $Z$ markers $(m_1, m_2,\dots,m_Z)$. Given event sequence $u$ with event marker $m_i=\{(m_{i,1},m_{i,2},\dots,m_{i,Z})\}$ for $m_{i,z}\in \{1,2,...,M_{z}\}$ where $z\in\{1,2,\dots,Z\}$, the intensity function of a conventional marked point process model for marker $m$ is defined by:
\begin{equation}\label{eq:coupled_int}
	\lambda_{m}^{u}(t) = f\left(\bm{\theta}_{m}^{\top}\bm{x}_{o}^{u}h(t) +\sum_{i:t_{i}^{u}<t}\bm{a}_{mm_i}g(t,t_{i})\right),
\end{equation}
where $\bm{x}_{0}^{u}\in \mathbb{R}^{M}$ is the time-invariant features of sequence taker $u$ extracted from its profile, like \emph{Self-introduction} of LinkedIn users or patients' \emph{diagnose} in MIMIC-II database, and $\bm{\theta}_m\in \mathbb{R}^{M}$ is the corresponding coefficients.

For the choice of the three functions $f$, $h$, $g$ in Eq. \ref{eq:coupled_int}, there are many forms in the literature that can be abstracted by the above framework, and popular ones are depicted in Table \ref{tab:time-function}.

For marked point process model, when the marker contains multiple label dimensions, one major bottleneck is that this model involves the infectivity matrix $\bm{a}$ with size $(\prod_{z=1}^{Z}M_z) \times (\prod_{z=1}^{Z}M_z)$ to measure the directional effect between $m_i$ and $m_j$. More generally, the size of the infectivity matrix $\bm{a}$ is $O(n^{2Z})$ (assume all dimensions $M_z$ have same number of values by $n$),
which incurs learning difficulty.

\begin{figure}[tb!]
	\centering
	\subfigure{\includegraphics[width=0.46\textwidth]{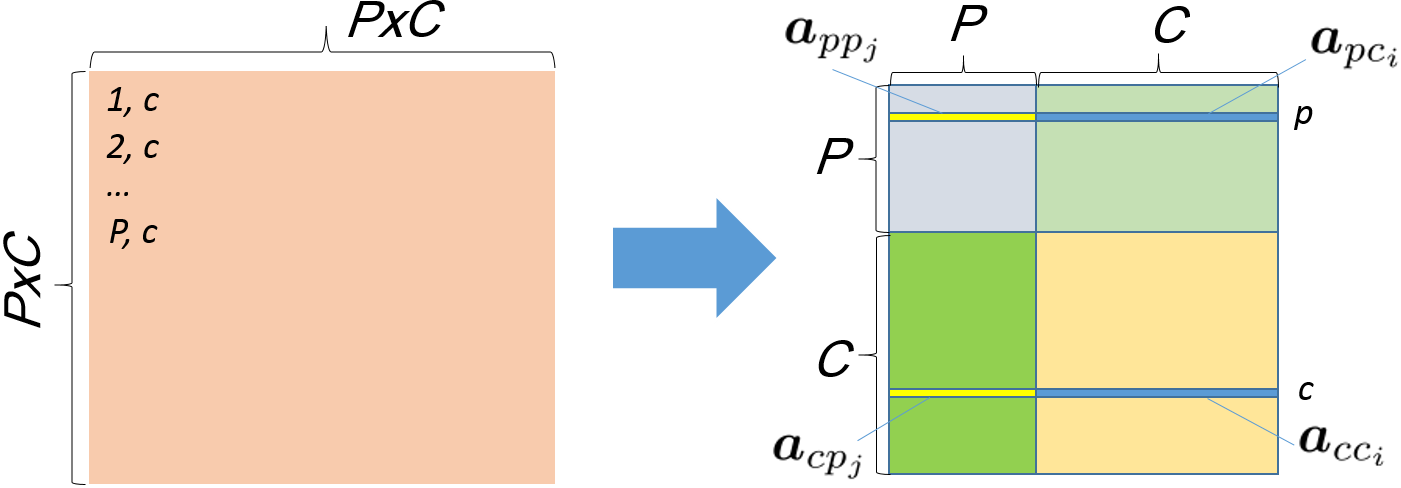}}
	\caption{Decoupling infectivity matrix between two markers $C$ and $P$ from size $\mathbb{R}^{(P\times C)\times(P\times C)}$ to $\mathbb{R}^{(P+C)\times(P+C)}$. Note the indicated rows for $\bm{a}_{pp_j}$, $\bm{a}_{pc_i}$, $\bm{a}_{cp_j}$, $\bm{a}_{cc_i}$.}
	\label{fig:matrix_a}
\end{figure}

To mitigate the challenge for learning the above parameter matrix, especially when $\prod_{z=1}^{Z}M_z$ is large while the sequences are relatively short, we propose the decoupled factorial point process model to linearly decouple the above intensity function into $Z$ interdependent point processes $\lambda_{m_z}^{u}(t),z\in\{1,2,\dots,Z\}$ for the $z$-th marker as written by:
\begin{small}
\begin{equation}\label{eq:intensity_c}
f\bigg(\underbrace{\bm{\theta}_{z}^{\top}\bm{x}_{o}^{u}h(t)}_\text{effect by profile}
 +\underbrace{\sum_{i:t_{i}^{u}<t}\bm{a}_{zz_{i}}\bm{b}^u_{z_{i}}g(t,t_{i})}_{\text{effect by former markers $z_i$}} + \underbrace{\sum_{\begin{subarray}{l}y=1\\y\neq z \end{subarray}}^{Z}\sum_{j:t_{j}^{u}<t}\bm{a}_{zy_{j}}\bm{b}^u_{y_{j}}g(t,t_{j})}_{\text{effect by former markers $y_i$}}\bigg)
\end{equation}
\end{small}
where $\bm{b}^u_{z_{i}}\in \{0,1\}^{M_z}$, $\bm{b}^u_{y_{j}}\in\{0,1\}^{M_y}$ is the binary indicators connecting through the influence of one's former marker $z_{i}$ and markers $y_{j}$. Note the row vector $\bm{a}_{zz_i}\in \mathbb{R}^{1\times M_z}$ is the parameter for intra-influence within marker $\{z_i\}$, and $\bm{a}_{zy_i}\in \mathbb{R}^{1\times M_y}$ is for inter-influence between $z_i$ and $y_i$. The above vectors are illustrated in Fig.\ref{fig:matrix_a} when $Z=2$, using notation $c$ as $z_1$ and $p$ as $z_2$.

In fact, function $f(\cdot)$, $g(\cdot)$, $h(\cdot)$ are predefined and some embodiments can be chosen from Table \ref{tab:time-function}. For the time being, we do not specify these functions while focus on solving the learning problem in a general setting.

\begin{figure*}[tb!]
	\centering
	\subfigure{\includegraphics[width=0.8\textwidth]{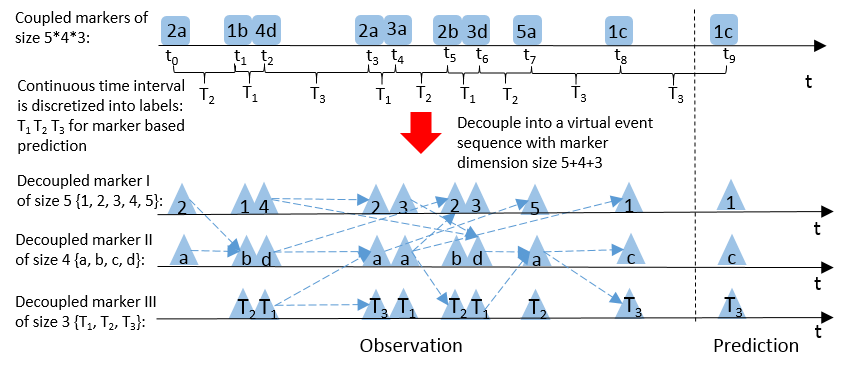}}
	\caption{Example of our decoupling perspective on factorial event sequence learning. The raw event sequence is represented with three-marker of label space $5\times 4 \times 3$. Our decouple model treats the raw sequence as an overlay of three sequences whose marker space is $5$, $4$, $3$ respectively and then the whole marker space's dimension is in linear: $5 + 4 + 3$. The directed dash arrows between events sketch the effect from previous events to future events: not only within one of the three sequences but also across the three sequences. Two particular attentions shall be paid to our model: 1) the method is designated to predict next event's marker but not for its continuous occurrence timestamp (see more details later in the paper). To enable next event time prediction, we discretize the time interval into several levels as illustrated by $\{T_2,T_1,T_3\}$. 2) On the other hand, the accurate timestamp $\{t_1,t_2,\dots\}$ rather than the discretized version $\{T_2,T_1,T_3\}$, is used for learning of the point process model which makes sure our model can capture the fine-grained raw time information.}
	\label{fig:example}
\end{figure*}

A concrete example is presented in Fig.\ref{fig:example}. See the caption for more details.

\subsection{Next-event marker prediction with regularizer}
\textbf{Loss function for discriminative learning}
Based on the defined intensity function, we write out the probability $P(m,t | \mathcal{H}_t^u)$ for event $m = \{m_1,m_2,\dots, m_Z\}$ happens at time $t$, conditioned on $u$'s history $\mathcal{H}_t^u$:
\begin{align}\notag
P(m, t | \mathcal{H}_t^u)
=&\lambda_{m}^u(t)\exp\left(-\sum_{m'}\int_{t_I^{u}}^{t}\lambda_{m'}^u(s)ds\right)\\\notag
=&\frac{\lambda_{m}^u(t)}{\sum\lambda_{m'}^u(t)}
\times\frac{\sum_{m'}\lambda_{m'}^u(t)}{\exp\left(\sum\int_{t_I^{u}}^{t}\lambda_{m'}^u(s)ds\right)}\\
=& P(m| t,\mathcal{H}_t^u)\times P(t|\mathcal{H}_t^{u}),
\end{align}
where $\lambda_{m}^u(t)$ is the event intensity, and $P(t|\mathcal{H}_t^u)$ is the conditional probability that this event happens at time $t$ given history $\mathcal{H}_t^u$. $P(m| t,\mathcal{H}_t^u)$ is the probability that the happened event is $m$ given current time $t$ and history $\mathcal{H}_t^u$.

Based on the above equation, most existing point process learning methods e.g. \cite{LewisJNS2011,ZhouICML13,DuNIPS15} fall into the generative learning framework aiming to maximize the joint probability of all observed events via a maximum likelihood estimator $\max_{\bm{\Theta}}\prod_{u,i} P(m_i^u, t_i^u|\mathcal{H}_{t_i^u}^u)$.

However, such an objective function is not tailored to the particular task at hand: instead of taking care of handling the posterior probability of the whole event sequence, we are more interested in predicting the next event and its mark information. To enable a more discriminative learning paradigm to boost the next event prediction accuracy, a recent work \cite{xu2016icu} suggests to focus on $P(m|t,\mathcal{H}_t^u)$ instead of $P(m, t | \mathcal{H}_t^u)$ as the loss function for learning.

In the decoupled point process model, the dependency between different markers has been measured by inter-influence parameters, i.e., the dependency between process $\lambda_{m_z}^{u}(t)$ for marker $m_z$ and process $\lambda_{m_y}^{u}(t)$ for marker $m_y$ have been measured by parameter $\bm{a}_{zy_{i}}$ and $\bm{a}_{yz_{i}}$ (see Eq. \ref{eq:intensity_c}) in an independent fashion. In the same spirit, here we simplify the probability $P(m| t,\mathcal{H}_t^u)$ by an independence assumption for marker $c$ and marker $p$:
\begin{align}\label{eq:prob2}
P(m| t,\mathcal{H}_t^u)=& P({m_1,m_2,\dots,m_Z}| t,\mathcal{H}_t^u)\\\notag
=& \prod_{z=1}^{Z}P(m_z| t,\mathcal{H}_t^u) = \prod_{z=1}^{Z} \frac{\lambda_{m_z}^u(t)}{\sum_{m'_{z}}\lambda_{m'_{z}}^u(t)} ,
\end{align}
where $P(m_z| t,\mathcal{H}_t^u)$ is the normalized intensity function. This simplification leads to the following loss:
\begin{align}\notag\label{cost}
L(\bm{\Theta})=&-\sum_{u=1}^{U}\sum_{i=1}^{N^u}\sum_{z=1}^{Z}
\sum_{m_z=1}^{M_z}1\{m_{i,z}^u=m_z\}\log P(m_z| t_{i-1}^u, \mathcal{H}_{t_{i-1}^u}^u)\\
=&-\sum_{u=1}^{U}\sum_{i=1}^{N^u}\log\left(\prod_{z=1}^{Z}\frac{\lambda_{m_{i,z}^u}^u(t_{i-1}^u)}{\sum_{m'_z}\lambda_{m'_z}^u(t_{i-1}^u)}\right),
\end{align}
where $1\{statement\}$ is an indicator returning 1 if true, otherwise 0. $\bm{\Theta}=\{\bm{\Theta_1};
\bm{\Theta_2};\dots;\bm{\Theta_z};\dots;\bm{\Theta_Z}\} \in \mathbb{R}^{(\sum_{z=1}^{Z}M_z)\times(M+\sum_{z=1}^{Z}M_z)}$ is the parameters whereby $\bm{\Theta_c} = [\bm{\theta_z}^{\top},\bm{a_{zz}},\bm{a_{zy}}|_{y=1,y\neq z}^{Z}] \in \mathbb{R}^{M_z\times(M+\sum_{z=1}^{Z}M_z)}$
\begin{table}[tb!]
	\centering
	\caption{Parametric forms of popular point processes.}
        \resizebox{0.48\textwidth}{!}{
		\begin{tabular}{rccc}
            \toprule
			Model  &$f(x)$  &$h(t)$  &$g(t,t')$    \\\midrule
			Modulated Poisson process (\textbf{MPP}) \cite{lloyd2014variational}    &$x$    &$1$    &$1$     \\
			Hawkes process (\textbf{HP}) \cite{LewisJNS2011}                     &$x$    &$1$    &$e^{-w(t-t')}$    \\
			Self-correcting process (\textbf{SCP}) \cite{isham1979self}        &$e^x$   &$t$     &$1$  \\
			Mutually-correcting process (\textbf{MCP}) \cite{xu2016icu}  &$e^x$ &$t-t_{I}$            &$e^{-\frac{(t-t')^2}{\sigma^2}}$  \\
			\bottomrule
		\end{tabular}}\label{tab:time-function}
\end{table}

\textbf{Sparse group regularization}
Since the model involves many parameters for learning, a natural idea is introducing sparsity to reduce the complexity. Incorporating both group sparsity and overall sparsity, we use a sparse group regularizer \cite{simon2013sparse} as the regularization, a combination of $\ell_1$ regularization $\|\bm{\Theta}\|_1$ and a group lasso $\|\bm{\Theta}\|_{1,2} = \sum_{j=1}^{M+\sum_{z=1}^{Z}M_z}\|\theta_j\|_2$. The group lasso encourages the nonzero elements concentrated on a few columns in the whole matrix $\bm{\Theta}$, and the rest part is assumed to be zeros. The $\ell_1$ regularization encourages the whole matrix $\bm{\Theta}$ to be sparse. The behind rationale is that only a few profile features and event marker values will be the main contributor to the point process. This means only a few columns will be activated. As a result, the regularized objective is:
\begin{eqnarray}\label{eqn:final-obj}
\min_{\bm{\Theta}}L(\bm{\Theta}) + \lambda_1\|\bm{\Theta}\|_1 + \lambda_2\sum_{j=1}^{N}\|\theta_j\|_2,
\end{eqnarray}
where $N=M+\sum_{z=1}^{Z}M_z$, $\bm{\Theta}\in\mathbb{R}^{(\sum_{z=1}^{Z}M_z)\times N}$, $\lambda_1 = \alpha\lambda$ and $\lambda_2 = (1-\alpha)\lambda$, $\lambda$ is the regularization weight, $\alpha\in (0,1)$ controls the balance between overall and group sparsity.

\section{Learning Algorithm}
In this section, we first present our tailored algorithm to the presented model. Then we give a new perspective and show how to reformulate it into a Logistic Regression task.

Following the scheme of ADMM, we propose a FISTA \cite{beck2009fast} based method with Line Search and two soft-shrinkage operators to solve the subproblems of ADMM. The whole algorithm is summarized in Alg. \ref{alg:ADMM}.
\subsection{Soft Shrinkage Operator}
First we review two soft-shrinkage operators \cite{donoho1995noising} that solve the following two basic minimization problem, which will be used in our algorithm.
\begin{itemize}
\item The minimization problem
$$\arg\min_{\bm{\theta}\in \mathbb{R}^p}\{\|\bm{\theta}\|_1 + \frac{u}{2}\|\bm{\theta}-\bm{r}\|^2\},$$
with $u>0$, $\bm{\theta} \in \mathbb{R}^p$, $\bm{r} \in \mathbb{R}^p$, has a closed-form solution given by the soft-shrinkage operator $\bm{\textbf{shrink}_{1,2}}$ defined:
$$\bm{\theta}^* = \text{shrink}_{1,2}(\bm{r}, 1/u) \triangleq \text{sign}(\bm{r}) \cdot \max\{0, |\bm{r}| - 1/u\},$$
where $\text{sign}(\cdot)$ is the sign function.
\item The minimization problem
$$\arg\min_{\bm{\theta}\in \mathbb{R}^p}\{\|\bm{\theta}\|_2 + \frac{u}{2}\|\bm{\theta} - \bm{r}\|^2\},$$
with $u>0$, $\bm{\theta} \in \mathbb{R}^p$, $\bm{r}\in \mathbb{R}^p$, has a closed-form solution given by the soft-shrinkage operator $\bm{\textbf{shrink}_{2,2}}$ defined:
$$\bm{\theta}^* = \text{shrink}_{2,2}(\bm{r}, 1/u) \triangleq \frac{\bm{r}}{\|\bm{r}\|_2} \cdot \max\{0, \|\bm{r}\|_2 - 1/u\}.$$
\end{itemize}
\subsection{ADMM iteration scheme}
To solve the minimization problem defined in Eq. \ref{eqn:final-obj} using ADMM solver, we add two auxiliary variables $\beta$, $\gamma$. The augmented Lagrangian function for Eq. \ref{eqn:final-obj} is $L_u^s(\bm{\Theta}, \bm{\beta}, \bm{\gamma})=$
\begin{equation}\label{equ:augLoss}
L(\bm{\Theta}) + \lambda_1\|\bm{\Theta}\|_1 + \lambda_2\sum_{j=1}^{N}\|\gamma_j\|_2 - \bm{\beta}^T(\bm{\Theta}-\bm{\gamma}) + \frac{u}{2}\|\bm{\Theta}-\bm{\gamma}\|_2^2,
\end{equation}
where $\bm{\beta} = (\beta_1, \beta_2, \dots, \beta_N)$, $\bm{\gamma} = (\gamma_1, \gamma_2, \dots, \gamma_N)$, $\bm{\Theta} = (\theta_1, \theta_2, \dots, \theta_N)$, and $\beta_j,\gamma_j,\theta_j\in \mathbb{R}^{C+P}$.

The iterative scheme is given by:
\begin{small}
\begin{align}
\label{eq:sub1}\bm{\Theta}^{k+1} &= \arg\min_{\bm{\Theta}} L_u^s(\bm{\Theta}, \bm{\beta}^k, \bm{\gamma}^k)\\\notag
    &=\arg\min_{\bm{\Theta}} \{L(\bm{\Theta}) - \bm{\beta}^T(\bm{\Theta}-\bm{\gamma}) + \frac{u}{2}\|\bm{\Theta}-\bm{\gamma}\|_2^2 + \lambda_1\|\bm{\Theta}\|_1\}\\
\label{eq:sub2}\bm{\gamma}^{k+1} &= \arg\min_{\bm{\gamma}} L_u^s(\bm{\Theta}^{k+1}, \bm{\beta}^k, \bm{\gamma}) \\\notag
    & = \arg\min_{\bm{\gamma}} \{\lambda_2\sum_{j=1}^N\|\gamma_j\|_2 - \bm{\beta}^T(\bm{\Theta}-\bm{\gamma}) + \frac{u}{2}\|\bm{\Theta}-\bm{\gamma}\|_2^2\}\\
\bm{\beta}^{k+1} &= \bm{\beta}^k - u(\bm{\Theta}^{k+1} - \bm{\gamma}^{k+1})
\end{align}
\end{small}

%

Therefore the optimization of Function \ref{equ:augLoss} has been divided into two sub-problems defined as Eq. \ref{eq:sub1} and Eq. \ref{eq:sub2}. While for Eq. \ref{eq:sub2}, the update of $\bm{\gamma}$, it has a closed-form solution given by operator $\bm{\textbf{shrink}_{2,2}}$ as follows
\begin{eqnarray}\label{eqn:updateGamma}
\begin{aligned}
\gamma_j^{k+1} & = \arg\min_{\gamma_j}\{\lambda_2\|\gamma_j\|_2 + \frac{u}{2}\|\gamma_j - (\bm{\Theta}^{k+1} - \bm{\beta}^k/u)_j\|_2^2\} \\
& = \textbf{shrink}_{2,2}((\bm{\Theta}^{k+1} - \bm{\beta}^{k}/u)_j, \lambda_2/u) \\
& = \nu_j^k - P_{\mathcal{D}_j}(\nu_j^k),
\end{aligned}
\end{eqnarray}
where $\nu_j^k = (\bm{\Theta}^{k+1} - \bm{\beta}^{k}/u)_j$, $\mathcal{D}_j$ denotes the ball in $p_j$-dimension centered at 0 with radius $\lambda_2 / u$ \cite{donoho1995noising}.
\subsection{FISTA with line search}
To solve Eq. \ref{eq:sub1} we define $g(\bm{\Theta}) = L(\bm{\Theta}) + \frac{u}{2}\|\bm{\Theta} - \bm{\gamma}^k - \bm{\beta}^k/u\|_2^2$,
then $\bm{\Theta}^{k+1}$ can be obtained by solving
$\label{eqn:FISTA}
\bm{\Theta} = \arg\min\limits_{\bm{\Theta}}\{g(\bm{\Theta}) +\lambda_1\|\bm{\Theta}\|_1 \}
$
through a FISTA method \cite{beck2009fast} with line search to compute the step size. The Algorithm is summarized in Alg.\ref{alg:FISTA}

\begin{algorithm}[tb!]
    \caption{\textbf{FISTA}($\bm{\Theta}^k$)}
    \label{alg:FISTA}
    \KwIn{$\bm{\Theta}^k$ from last iteration}
    Initialize $\bm{\Theta}^{(k,0)} = \bm{v}^{(0)} = \bm{\Theta}^k$, threshold $\epsilon=0.01$, $i=1$,$\tau_i = \frac{2}{i+1}$,$t_0 = \hat{t}>0$, $\eta=0.8$ \;
    \While {$\frac{\|\bm{\Theta}^{(k,i)}-\bm{\Theta}^{(k,i-1)}\|_2}{\|\bm{\Theta}^{(k,i)}\|_2}\leq\epsilon$}
    {
        $\bm{y}=(1-\tau_k)\bm{\Theta}^{(k,i-1)} + \tau_k \bm{v}^{(i-1)}$, $t=t_{i-1}$\;

        $\bm{\Theta}^{(k,i)} = \textbf{shrink}_{1,2}(\bm{y}-\frac{1}{t} \nabla g(\bm{y}), \lambda_1/t)$\;
        \While {$g(\bm{\Theta^{(k,i)}}) > g(\bm{y}) + \nabla g(\bm{y})^T (\bm{\Theta}^{(k,i)} - \bm{y}) + \frac{1}{2t}\|\bm{\Theta}^{(k,i)} - \bm{y}\|_2^2$}
        {
            $t = \eta\cdot t $\;
            $\bm{\Theta}^{(k,i)} = \textbf{shrink}_{1,2}(\bm{y}-\frac{1}{t} \nabla g(\bm{y}), \lambda_1/t)$\;
        }
        $\bm{v}^{(i)} = \bm{\Theta}^{(k,i-1)} + \frac{1}{\tau_k}(\bm{\Theta}^{(k,i)} - \bm{\Theta}^{(k, i-1)})$, $i=i+1$\;

    }
    $\bm{\Theta}^{k+1} = \bm{\Theta}^{(k,i)}$, \Return $\bm{\Theta}^{k+1}$\;
\end{algorithm}
\begin{algorithm}[tb!]
    \caption{\small{Decoupled Learning of Factorial Point Process}}
    \label{alg:ADMM}
    \KwIn{ two associated marked point process $\{(c_i,p_i,t_i)\}$, $\lambda_1 > 0$,$\lambda_2 > 0$, threshold $\epsilon = 0.01$}
    Initialize $(\bm{\Theta}, \bm{\beta}, \bm{\gamma}) = (\bm{0},\bm{0},\bm{0})$\;
    \While {$\frac{\|\bm{\Theta}^k-\bm{\Theta}^{k-1}\|_2}{\|\bm{\Theta}^k\|_2}\leq\epsilon$}
    {
        Update $\bm{\Theta}^{k+1}$ via $\bm{\Theta}^{k+1} = \textbf{FISTA}(\bm{\Theta}^k)$\;
        Compute $\bm{\gamma}^{k+1}$ via Eq. \ref{eqn:updateGamma};
        Update $\bm{\beta}^{k+1}$ via $\bm{\beta}^{k+1} = \bm{\beta}^k - u(\bm{\Theta}^{k+1} - \bm{\gamma}^{k+1})$\;
        $k=k+1$\;
    }
    \KwOut {$\bm{\Theta}^k$}
\end{algorithm}

\subsection{Reformulating to Logistic Regression task}\label{sec:reformulate}
Based on the intensity function Eq. \ref{eq:intensity_c}, the loss function Eq. \ref{cost}, we show how to reformulate the learning of the decoupled point process as a multi-class Logistic Regression task. One obvious merit of this reformulation is the reuse of on-the-shelf LR solvers e.g. http://www.yelab.net/software/SLEP/ \cite{Liu:2009:SLEP:manual} with little parameter tuning. In contrast, the algorithm presented in Alg.\ref{alg:ADMM} involves more parameters and is more computational costive as shown in Fig.\ref{fig:Convergence} and Table \ref{tab:raw&reformulated_LinkedIn}.

For event taker $u$ at time $t$, by separating the event taker $u$'s feature $\bm{f_t^{u}} = [x_0^uh(t), \sum_{i:t_i^u<t}b_{z_i}^ug(t,t_i), \sum_{y=1,y\neq z}^{Z}\sum_{j:t_j^u}b_{y_{j}}^ug(t,t_j)]\in \mathbb{R}^{M+\sum_{z=1}^{Z}M_z}$ from the parameters $\bm{\Theta}_z$, the conditional intensity function in Eq. \ref{eq:intensity_c} can be written as:
\begin{equation}
\lambda_{m_z}^{u}(t) = \exp(\bm{\Theta}_z\bm{f_t^{u}})
\end{equation}
Therefore the probability $P(p| t,\mathcal{H}_t^u)$ can be written as:
\begin{equation}\label{eq:c-prob}
P(m_z| t,\mathcal{H}_t^u) = \frac{\exp(\bm{\Theta}_z\bm{f_t^{u}})}{\sum_{z'}\exp(\bm{\Theta}_{z'}\bm{f_t^{u}})}.
\end{equation}
This is exactly the same probability function of sample $\bm{f_t^{u}}$ belonging to class $m_z$ for a Softmax classifier of $M_z$ classes, and $\bm{\Theta}_z$ is the parameter.

Hence the log-loss function in Eq. \ref{cost} becomes:
\begin{equation}\notag
L(\bm{\Theta})=-\sum_{u=1}^{U}\sum_{i=1}^{N^u}\log\left(
\prod_{z=1}^{Z}\frac{\exp(\bm{\Theta}_z\bm{f_{t_i}^{u}})}{\sum_{z'}\exp(\bm{\Theta}_{z'}\bm{f_{t_i}^{u}})}
\right),
\end{equation}
which is the sum of $Z$ Softmax classifiers' loss functions.

So far we have reformulated the decoupled learning of the factorial marked point process to the learning of $Z$ Softmax classifiers. For the $z$-th classifier, it takes $\bm{f_t^{u}}$ from the sample and classify it to one of $M_z$ markers $\hat{m_z}$. In the following experiments, we will show that the reformulated learning method in fact optimizes the same loss function as Alg.\ref{alg:ADMM}.

\subsection{Event marker prediction}
After learning parameters $\bm{\Theta}=\{\bm{\Theta}_z|_{z=1}^{Z}\}$, we can predict the next event markers $m=\{\hat{m_z}|_{z=1}^{Z}\}$ at $t$, given history $\mathcal{H}_t^u$ by computing $P(m_z| t,\mathcal{H}_t^u)$ (see Eq. \ref{eq:c-prob}). The predictions $\hat{c}$ and $\hat{m_z}$ are given by $\hat{m_z} = \arg\max_{\hat{m_z}\in M_z} P(m_z| t,\mathcal{H}_t^u)$. It is important to note that though our model technically only issues discrete output as it is inherently a classification model, while in practice the future events' timestamp can be predicted by an approximated discrete duration as done in our experiments. In this regard, we treat the future timestamp as a marker. 


\section{Empirical Study and Discussion}\label{Experiments}
\subsection{Dataset and protocol}
To verify the potential of the proposed model, we apply it to a \emph{LinkedIn-Career} dataset crawled and de-identified from LinkedIn to predict user's next company $\hat{c}$, next position $\hat{p}$ and duration $\hat{t}$ of current job; an ICU dataset extracted from public medical database MIMIC-II \cite{goldberger2000physiobank} to predict patient's transition to the next ICU department $\hat{c}$ and duration of stay $\hat{p}$ in current department. Experiments are conducted under Ubuntu 64bit 16.04LTS, with i7-5557U 3.10GHz$\times$4 CPU and 8G RAM. For the convenience of replicating the experiments, the crawled \emph{de-identified LinkedIn-Career} dataset and the code is available on Github\footnote{https://github.com/blade091shenwei/factorial-marked-point-process}.

\textbf{Dataset}
The \emph{LinkedIn-Career} Dataset contains $5,006$ users crawled from \emph{information technology} (IT) industry on LinkedIn (\emph{https://www.linkedin.com/}), including their \emph{Self-introduction}, \emph{Technical skills} and \emph{Working Experience} after de-identification preprocess. We collect samples in IT industry because: i) The staff turnover rate is high, which makes it easier to collect suitable samples; ii) The IT industry is most familiar to the authors, and our domain knowledge can help better curate the raw data. We extract profile features from users' \emph{Self-introduction} and \emph{Technical skills}, and get users' history company and position $\{(c_i,p_i)\}$ from \emph{Working Experience}. After we exclude samples with zero job movement, we have a so-called \emph{LinkedIn-Career} benchmark, involving 2,403 users, 57 IT companies, 10 kinds of positions and 4 kinds of durations. The dataset is to some extent representative for IT industry. For companies, we have large corporations like \emph{Google}, \emph{Facebook}, \emph{Microsoft} and medium-sized enterprise like \emph{Adobe}, \emph{Hulu}, \emph{VMWare}. For positions we have technical positions like \emph{engineer}, \emph{senior engineer}, \emph{tech lead}, and management positions like \emph{manager}, \emph{director}, \emph{CEO}. For durations we discretize the duration of stay in a position or company as \emph{temporary}( within 1 year), \emph{short-term}( 1-2 years), \emph{medium-term}( 2-3 years) and \emph{long-term}( more than 3 years). The goal is to predict user's next company $\hat{c}$ from $C=57$ companies, next position $\hat{p}$ from $P=10$ positions and duration of stay in current company and position $\hat{t}$ from $T=4$ durations. 

The \emph{ICU} dataset contains $30,685$ patients from MIMIC-II database, including patients' \emph{diagnose}, \emph{treatment record}, \emph{transition} between different ICU departments and \emph{duration} of stay in the departments. The goal is to predict patient's next ICU department $\hat{c}$ from $C=8$ departments including Coronary care unit (CC), Anesthesia care unit (ACU), Fetal ICU (FICU), Cardiac surgery recovery unit (CSRU), Medical ICU (MICU), Trauma Surgical ICU (TSICU), Neonatal ICU (NICU), and General Ward (GW), and predict patient's duration of stay $\hat{t}$ from $T=3$ kinds of duration including \emph{temporary}( within 1 day), \emph{short-term}( 1-5 days), and \emph{long-term} (more than 5 days). The profile features are extracted from patients' \emph{diagnose} (ICD9 code of patients' disease) and \emph{treatment record} (nursing, medication, treatment).

Many peer methods are evaluated as follows:

\textbf{Intensity function choices}
Our framework is tested by four point process embodiments namely: i) Mutually-corrected Processes (\textbf{MCP}), ii) Hawkes Process (\textbf{HP}), iii) Self-correcting Process (\textbf{SCP}) and iv) Modulated Poisson Process (\textbf{MPP}). Their characters are briefly compared in Table \ref{tab:time-function}. Note in our experiments, all these models are learned via the reformulated LR algorithm as described in Alg. \ref{alg:ADMM}.

\textbf{Comparison to classic Logistic Regression} We test a non-point process approach i.e. the plain LR. For the point process based LR solver, its input is $\bm{f_t^{u}} = [x_0^uh(t), \sum_{i:t_i^u<t}b_{z_i}^ug(t,t_i), \sum_{y=1,y\neq z}^{Z}\sum_{j:t_j^u}b_{y_{j}}^ug(t,t_j)]$, while the plain LR involves the raw feature as $\bm{f^u} = [x_0^u, b_{z_I}^u, \sum_{y=1,y\neq z}^{Z}b_{y_{I}}^u]$, including user profile feature $x_0^u$, binary indicator $b_{z_I}^u$ and $\sum_{y=1,y\neq z}^{Z}b_{y_{I}}^u$ representing one's current state without considering the history states.

\textbf{Comparison to RNN and RMTPP} We also experiment on RNN by treating the prediction task as a sequence classification problem. A dynamic RNN that can compute over sequences with variable length is implemented.

Moreover, to explore the effect of discretizing the time interval when making duration prediction, we also experiment on RMTPP (Recurrent Marked Temporal Point Process) proposed by \cite{du2016recurrent}. Instead of predicting a discrete label for duration, it gives a continuous prediction result.

\textbf{Prediction performance metrics} We use prediction accuracy $\mbox{AC}$ to evaluate the performance of the model with four variants $\mbox{AC}_{c}$, $\mbox{AC}_{p}$, $\mbox{AC}_{t}$, $\mbox{AC}_{cpt}$ to denote the prediction accuracy for state $c$ (i.e. \# correct ones out of total predictions), state $p$, state $t$ and joint $c$,$p$,$t$ respectively.

To evaluate the performance of our discrete duration prediction compared with RMTPP, both MSE (Mean Squared Error) and AC are computed. To compute prediction MSE, the predicted discrete duration is substituted by the intermediate time point of the discrete intervals, e.g., 0.5 years for \emph{temporary} stay, 1.5 years for \emph{short-term} stay and 4 years for \emph{long-term} stay. To compute prediction AC, the predicted continuous duration of RMTPP is discretized using the same criterion by the proposed model.

\begin{figure*}[tb!]
	\centering
	\subfigure[Company]{
		\includegraphics[width=0.23\linewidth]{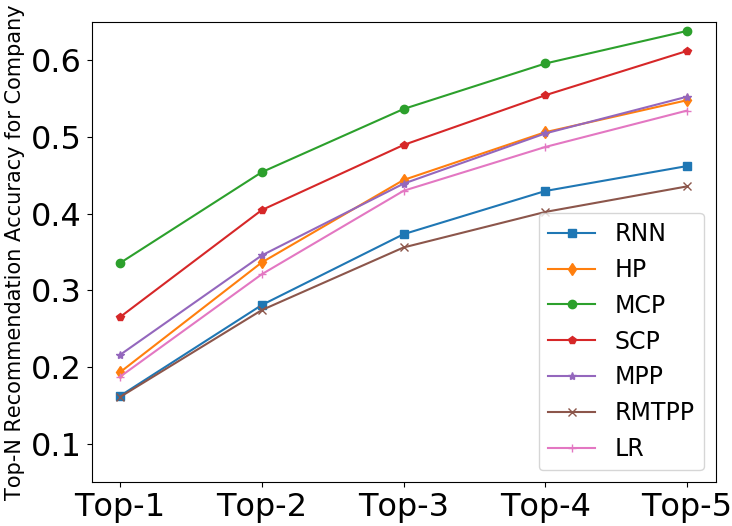}\label{fig:company}
	}
	\subfigure[Position]{
		\includegraphics[width=0.23\linewidth]{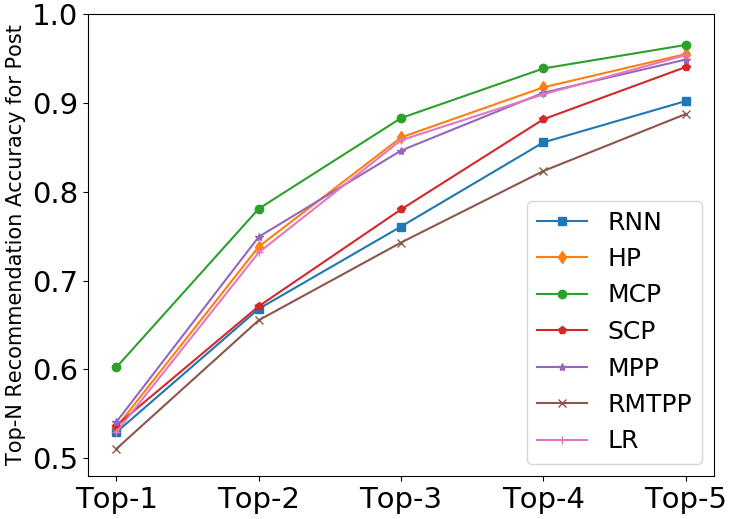}\label{fig:position}
	}
	\subfigure[Duration]{
		\includegraphics[width=0.23\linewidth]{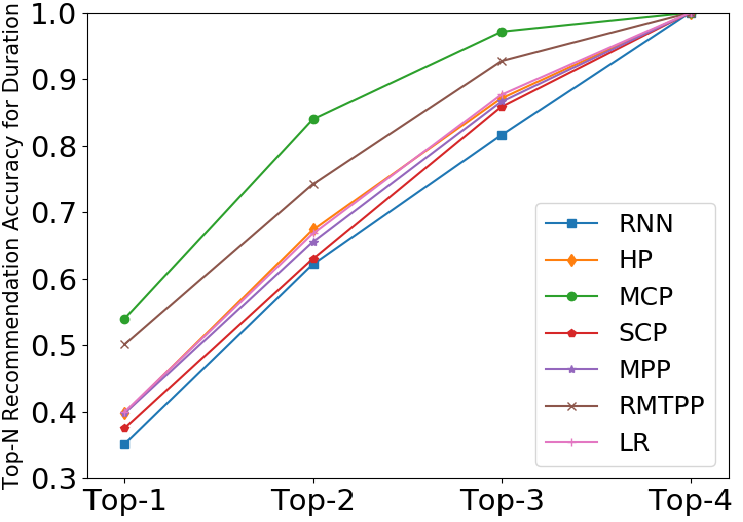}\label{fig:duration}
	}
    \subfigure[Company \& Position \& Duration]{
		\includegraphics[width=0.23\linewidth]{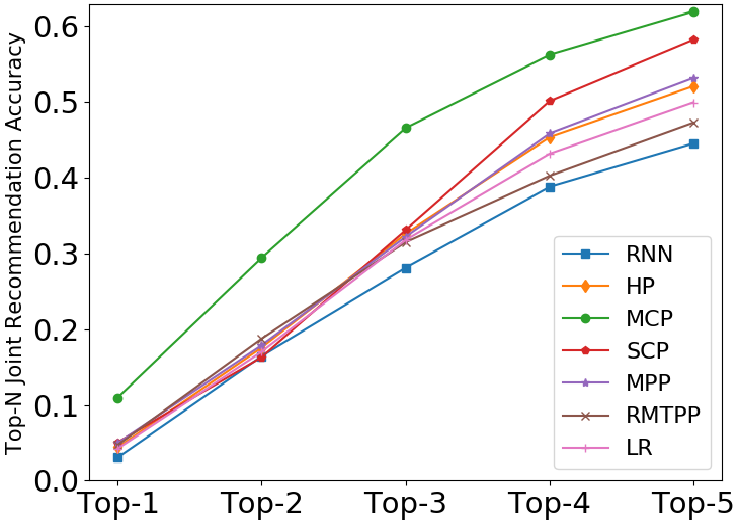}\label{fig:joint}
	}
	\caption{Top-$5$ prediction accuracy on the collected \emph{LinkedIn-Career} dataset out of 57 companies, 10 positions and 4 durations.}\label{fig:Top-N}
\end{figure*}

For \emph{LinkedIn-Career} data, we further compute precision curve for the top-K position, company and duration predictions as shown in Fig. \ref{fig:Top-N}. These metrics are widely used for recommender system. In fact, as our model is for predicting the next company $\hat{c}$, next position $\hat{p}$ and duration $\hat{t}$ given career history $\mathcal{H}_t^u$, it can be used for recommending companies and posts at the predicted time period $\hat{t}$.

All the experimental results are obtained by 10-fold cross validation as commonly adopted like \cite{YanIJCAI13}.
\begin{figure}[tb!]
	\centering
	\subfigure{
		\includegraphics[width=0.45\linewidth]{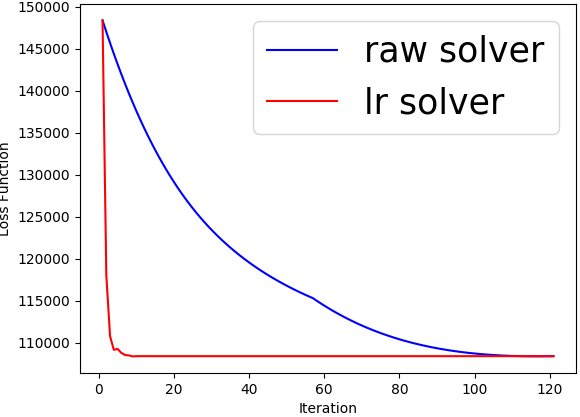}\label{fig:convergence_LinkedIn}
	}
	\subfigure{
		\includegraphics[width=0.45\linewidth]{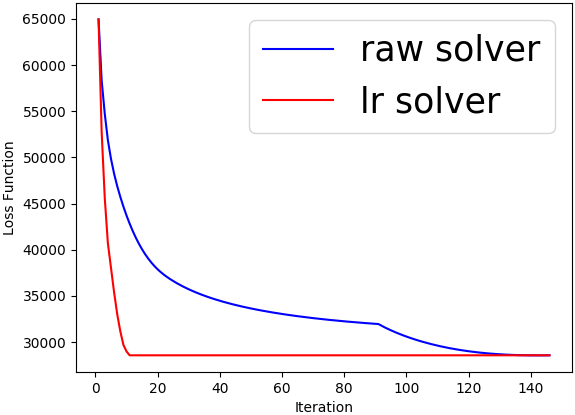}\label{fig:convergence_ICU}
	}
	\caption{Convergence curve of reformulated LR solver \& ADMM solver (Alg.\ref{alg:ADMM}) by similar random initialization. Left: \emph{ICU} dataset from MIMIC-II; Right: \emph{LinkedIn-Career}.}\label{fig:Convergence}
\end{figure}

\begin{table}[tb!]
	\centering
	\caption{Comparison of the raw ADMM solver (Alg.\ref{alg:ADMM}) and the reformulated LR solver: prediction accuracy by percentage for $AC_c$, $AC_p$, $AC_t$, joint prediction accuracy $AC_{cpt}$, time cost and average iteration count by random initialization for 10 trials. Time and iteration number is the average result.}
	\resizebox{0.5\textwidth}{!}{
		\begin{tabular}{r|r|ccccrr}
			\toprule
            Dataset& Method & $AC_c$ & $AC_p$ & $AC_t$ & $AC_{cpt}$& Time& Iter. \#\\ \midrule
			\multirow{2}{*}{Career} & Alg.\ref{alg:ADMM} & 32.81 & 60.67 & 52.41 & 10.74 & 123.8m & 147.7  \\
									&LR 				 & 33.58 & 60.13 & 53.96 & 10.96 & 46.8s & 11.2 \\ \midrule
            \multirow{2}{*}{ICU} &  Alg.\ref{alg:ADMM} & 76.63 & --- & 55.74 & 45.64 & 764.2m
             & 121.5 \\
								 &LR 				   & 76.98 & --- & 55.63 & 45.55 & 55.1s & 9.7 \\
        \bottomrule
		\end{tabular}}\label{tab:raw&reformulated_LinkedIn}
\end{table}

\begin{table*}[ht!]
\centering
\caption{Accuracy comparison for different intensity functions on \emph{LinkedIn-Career} (HP, SCP, MPP, MCP, see Table \ref{tab:time-function}). Numbers in bold denote the best or second-best accuracy on the specified metric and dataset. Learning for all point process based models is via the reformulated LR solver as discussed in the main paper. Long sequence denotes those with more than 2 job transitions. For the non-point process based (classic) LR$_{np}$, we present its performance for each prediction target.}
\resizebox{1\textwidth}{!}{
	\begin{tabular}{r|l|ccc|ccc|ccc|ccc}
	\toprule
	\multicolumn{2}{c|}{}&\multicolumn{3}{c}{company pred. accuracy $AC_c(\%)$}&\multicolumn{3}{|c}{position pred. accuracy $AC_p(\%)$} & \multicolumn{3}{|c}{duration pred. accuracy $AC_t(\%)$} & \multicolumn{3}{|c}{joint pred. accuracy $AC_{cpt}(\%)$} \\ \hline
	sequence &intensity & decoupled & coupled & uni-com & decoupled & coupled & uni-pos & decoupled & coupled & uni-dur & decoupled & coupled & uni-cpt\\ \midrule
	\multirow{6}{*}{short}
	&HP & 15.37 		 & 14.23 & 14.50 & 56.99 		  & 56.99 & 56.99 & 38.97 & 36.61 & 36.61 & 4.59 		  & 4.05 & 4.26 \\
	&SCP& 17.99 		 & 10.01 & 12.61 & 58.98 		  & 49.75 & 55.38 & 43.56 & 41.81 & 38.22 & 4.96 		  & 2.69 & 2.11 \\
	&MPP& 15.58 		 & 13.58 & 13.96 & 56.99 		  & 56.99 & 56.99 & 40.16 & 36.61 & 36.61 & 4.71 		  & 3.72 & 4.21 \\
	&MCP& \textbf{18.54} & 11.04 & 13.14 & \textbf{59.42} & 51.49 & 56.90 & \textbf{46.49} & 46.46 & 41.32 & \textbf{5.41} & 2.52 & 3.26 \\
	&LR$_{np}$ & --- 	 & 15.35 & --- 	 &  --- 		  & 56.99 & ---   & ---   & 38.31 & ---   &  --- 		  & 4.46 & ---\\
	&RNN & --- 	 & 13.18 & ---   & --- 			  & 56.98 & ---   & ---   & 36.94 & ---   & --- 		  & 3.77 & --- \\
	&RMTPP & --- 	 & 12.02 & ---   & --- 			  & 56.03 & ---   & ---   & 44.95 & ---   & --- 		  & 4.15 & --- \\ \midrule
	\multirow{6}{*}{long}
	&HP & 24.45 		 & 20.57 & 25.61 & 50.80 		  & 49.25 & 49.51 & 34.52 		   & 33.61 & 34.33 & 4.36 			& 3.53 & 4.32 \\
	&SCP& 35.41 		 & 20.49 & 26.46 & 51.00 		  & 40.54 & 46.96 & 33.27 		   & 28.99 & 28.64 & 7.47 			& 5.46 & 5.84 \\
	&MPP& 28.71 		 & 21.44 & 29.57 & 51.64 		  & 49.02 & 52.90 & 36.04 		   & 33.20 & 33.76 & 6.37 			& 4.17 & 6.49 \\
	&MCP& \textbf{50.23} & 29.98 & 44.12 & \textbf{60.45} & 50.16 & 55.65 & \textbf{47.00} & 40.33 & 37.78 & \textbf{14.61} & 7.21 & 10.05 \\
	&LR$_{np}$ & --- 	 & 24.59 &  ---  & --- 			  & 49.51 &  ---  &  --- 		   & 35.47 & ---   & --- & 4.14 & --- \\
	&RNN & --- 	 & 19.92 & ---   & --- 			  & 49.24 & ---   & --- 		   & 33.80 & ---   & --- & 3.26 & --- \\
	&RMTPP & --- 	 & 19.88 & ---   & --- 			  & 49.17 & ---   & --- 		  & 44.07 & ---   & --- 		   & 4.58 & --- \\ \midrule
	\multirow{6}{*}{all}
	&HP & 19.33 		 & 16.23 & 19.15 & 53.50 		  & 52.95 & 52.95 & 39.85 		  & 37.72 & 37.80 & 4.39 		   & 3.20  & 4.05 \\
	&SCP& 26.52 		 & 15.54 & 20.77 & 53.68 		  & 46.88 & 51.90 & 37.56 		  & 32.30 & 29.45 & 5.06 		   & 3.32  & 3.70 \\
	&MPP& 21.59 		 & 16.49 & 21.73 & 54.10 		  & 52.77 & 54.49 & 39.81 		  & 35.17 & 35.72 & 4.96 		   & 3.50  & 4.81 \\
	&MCP& \textbf{33.58} & 20.30 & 27.80 & \textbf{60.13} & 52.48 & 57.56 &\textbf{53.96} & 48.34 & 45.16 & \textbf{10.96} & 5.44  & 6.96 \\
	&LR$_{np}$ & --- 	 & 18.74 & ---   & --- 			  & 52.98 & ---   & --- 		  & 40.01 & ---   & --- 		   & 4.21  & --- \\
	&RNN & --- 	 & 16.24 & ---   & --- 			  & 52.93 & ---   & --- 		  & 35.21 & ---   & --- 		   & 3.02  & --- \\
	&RMTPP & --- 	 & 16.10 & ---   & --- 			  & 51.07 & ---   & --- 		  & 50.16 & ---   & --- 		   & 4.70 & --- \\\bottomrule
	\end{tabular}}\label{tab:results_LinkedIn}
\end{table*}
\begin{table*}[tb!]
	\centering
	\caption{Accuracy comparison for different intensity function models on ICU dataset from MIMIC-II.}
	\resizebox{1\textwidth}{!}{
	\begin{tabular}{r|l|ccc|ccc|ccc}
		\toprule
		\multicolumn{2}{c|}{}&\multicolumn{3}{c}{duration prediction accuracy $AC_t(\%)$}&\multicolumn{3}{|c}{transition prediction accuracy $AC_c(\%)$} & \multicolumn{3}{|c}{joint prediction accuracy $AC_{ct}(\%)$} \\ \hline
		sequence &intensity & decoupled & coupled & uni-duration & decoupled & coupled & uni-transition & decoupled & coupled & uni-dt\\ \midrule
		\multirow{6}{*}{all}
		&HP & 52.48 & 51.64 & 52.91 & 74.61 & 73.31 & 73.77 & 42.85 & 41.62 & 42.40 \\ 
		&SCP& 50.14 & 49.77 & 49.05 & 74.22 & 74.01 & 70.75 & 41.04 & 40.77 & 39.48 \\ 
		&MPP& 53.27 & 51.88 & 52.02 & 74.74 & 73.42 & 72.05 & 43.64 & 42.37 & 43.28 \\ 
		&MCP& \textbf{55.63} & 54.62 & 50.14 & \textbf{76.98} & 76.58 & 74.02 & \textbf{45.55} & 45.32 & 44.89 \\
		&LR$_{np}$ & --- & 39.88 & ---  & ---& 69.61 & --- & --- & 31.13 &--- \\
		&RNN & --- & 47.01 & ---  & --- & 70.54 & --- & --- & 36.44 &--- \\
		&RMTPP & --- & 54.28 & ---  & --- & 67.49 & --- & --- & 41.93 &--- \\ \bottomrule
	\end{tabular}}\label{tab:results_ICU}
\end{table*}

\subsection{Results and discussion}
We are particularly interested in analyzing the following main bullets via empirical studies and quantitative results.

\textbf{i) LR solver vs. ADMM solver }
To make a fair comparison, the LR solver and ADMM solver i.e. Alg.\ref{alg:ADMM} share the same initial parameter that initialized by a uniform distribution sampling from $(-1,1)$, and the running time and iteration count are the average of 10-fold cross validation.
Table \ref{tab:raw&reformulated_LinkedIn} compares LR solver and ADMM solver regarding with accuracy and time cost on the Dataset \emph{LinkedIn-Career} and \emph{ICU}. One can find the prediction accuracy is similar while the ADMM solver is more costive as we find it converges more slowly as shown in Fig.\ref{fig:Convergence}. Also, as shown in Alg.\ref{alg:ADMM}, it involves more hyper-parameters to tune and they have been tuned to their best performance.
For comparison between the reformulated LR via the point process framework, and the raw LR using only user profile i.e. LR$_{np}$, we find the former outperforms in most cases in Table \ref{tab:results_LinkedIn} and \ref{tab:results_ICU}.

Comparing running time in Table \ref{tab:raw&reformulated_LinkedIn}, the LR solver has better scalability than ADMM solver. This is because the ADMM solver is a general algorithm for convex optimization with sparse group regularization, while the LR solver works by a special design for the objectives that can be reformulated to Logistic Regression loss. Many algorithmic optimizations for Logistic Regression can be used in LR solver, like Efficient Projection in SLEP \cite{Liu:2009:SLEP:manual}.

\textbf{ii) Decoupled learning vs. RNN} As shown in Table \ref{tab:results_LinkedIn} and Table \ref{tab:results_ICU}, the decoupled marked point process model has much better performance than RNN. This is because the next-event prediction task for relatively short sequences like dataset \emph{LinkedIn-Career} and \emph{ICU} is not a typical sequence classification task. We need to make prediction on every step of the sequence, rather than make prediction at the end of the whole sequence. That means for the end-to-end RNN sequence classification model, it needs to deal with sequences with considerably variable length, including a large number of sequences of length $1$.

We also compare the accuracy of RMTPP that makes continuous duration prediction, with decoupled model and general RNN in Table \ref{tab:results_LinkedIn} and Table \ref{tab:results_ICU}. Though the duration prediction accuracy of RMTPP is improved compared to general RNN, the decoupled model still have better performance.

To further verify the effect of discretizing the time interval when making duration prediction, the MSE of decoupled-MCP and RMTPP is also compared in Table \ref{tab:mse}. Though it is a little tricky that discrete wrong prediction leads to smaller MSE than continuous wrong prediction, e.g., if a ground truth \emph{medium-term} duration of 2.5 years is misclassified as \emph{long-term} duration of 4 years for de-MCP, the RMTPP may gives a continuous prediction value of 10 years, the MSE of de-MCP is not only relatively but also absolutely small. It shows that the
discretization of time interval is to some extent rational.

\textbf{iii) Infectivity matrix decoupling vs. coupling}
Table \ref{tab:results_LinkedIn} and Table \ref{tab:results_ICU} also compare the performance of our decoupled model (see Eq. \ref{eq:intensity_c}) against the raw coupled model (see Eq. \ref{eq:coupled_int}), and the simplified model (single-dimension) when only marker $c$, $p$ or $t$ is considered. This boils down to the single-dimension case and the method is termed by uni-com, uni-pos and uni-dur for dataset \emph{LinkedIn-Career}, and uni-duration and uni-transition for \emph{ICU} respectively. While for uni-cpt, it involves no new model while uses the output of uni-c, uni-p and uni-t to combine them together as the joint prediction. It shows that the decoupled model consistently achieves the best performance, which perhaps is attributed to the reduction of model complexity given relatively limited training data.

Comparing the accuracies in Table \ref{tab:results_LinkedIn} for dataset \emph{LinkedIn-Career} and Table \ref{tab:results_ICU} for dataset \emph{ICU}, we can see that the improvement in accuracy of decoupled model compared to coupled model or single-dimension model, is more remarkable for \emph{LinkedIn-Career} than that for \emph{ICU}. The reason is that for \emph{LinkedIn-Career}, the coupled state space is decoupled from $C\times P\times T = 2280$ to $C+P+T=71$, and for \emph{ICU} it is decoupled from $C\times T=24$ to $C+T=11$. The \emph{LinkedIn-Career} dataset has a larger coupled state space than \emph{ICU}. So when decoupled to smaller state spaces, the improvement for \emph{LinkedIn-Career} is more notable than that for \emph{ICU}.

\textbf{iv) Choice of intensity function}
There are many popular intensity forms and some are listed in Table \ref{tab:time-function}. According to Table \ref{tab:results_LinkedIn} and Table \ref{tab:results_ICU}, the mutually-correcting process (MCP) consistently shows superior performance against other intensity function embodiments. This verifies two simple assumptions that i) the intensity tends to decrease for the moment the event happens, i.e., the desire for new job can be suppressed when a new job is fulfilled for job prediction, and patients' demand for transition to next ICU department decrease after they move into a new department for ICU department transition prediction; ii) the probability of future events is influenced by the history events according to Table \ref{tab:time-function}, i.e., one's transition possibility to new job is related to his/her history career experience, and patient's future ICU department transition procedure is related to his/her history treatment.

\textbf{v) Influence of sequence length}
To further explore the performance behavior, we experiment on short-sequences and long-sequences respectively on \emph{LinkedIn-Career}\footnote{ICU data is not included in the length test because $96\%$ of the patients in ICU have no more than three transitions.}. Results in Table \ref{tab:results_LinkedIn} show that the decoupled MCP algorithm has more advantage in long-sequence prediction, suggesting that the decoupled MCP can make better use of history information.

\begin{table}[tb]
	\centering
	\caption{MSE comparison of future event duration prediction for RMTPP and decoupled MCP on LinkedIn (in year) and ICU (in day) dataset. Note RMTPP model predicts continuous timestamp value for future events.}
	\resizebox{0.45\textwidth}{!}{
		\begin{tabular}{r|cc|cc}
			\toprule
			dataset &\multicolumn{2}{c|}{LinkedIn} &\multicolumn{2}{c}{ICU}\\ \midrule
			model & RMTPP & de-MCP & RMTPP & de-MCP \\
			MSE & 9.625 & 2.934 & 14.602 & 4.272 \\ \bottomrule
	\end{tabular}}\label{tab:mse}
\end{table}
\begin{table}[!tb]
	\centering
	\caption{Accuracy$(\%)$ by different regularizers. Sparse group (Eq. \ref{eqn:final-obj}) combines $\ell_1$ regularizer and group lasso.}
	\resizebox{0.48\textwidth}{!}{
		\begin{tabular}{r|r|ccc}
			\toprule
			dataset &marker &w/o sparse &group lasso &sparse group \\ \midrule
			\multirow{4}{*}{Career} &company & 29.16 & 31.47 &\textbf{33.58} \\
			&position & 56.99 & 58.16 &\textbf{60.13} \\
			&duration & 50.56 & 52.33 &\textbf{53.96} \\
			&joint (3) & 9.53 & 10.04 &\textbf{10.96} \\ \midrule
			\multirow{3}{*}{ICU} &duration &52.68 &55.13 &\textbf{55.63} \\
			&transition &73.45 &76.09 &\textbf{76.98} \\
			&joint (2)  &42.20 &45.49 &\textbf{45.55} \\ \bottomrule
		\end{tabular}}\label{tab:results_sparsity}
	\end{table}

\textbf{vi) Influence of sparsity}
To verify the effect of sparse group regularization, we compare the accuracy of the decoupled MCP model with different regularization settings, including \textbf{without sparse regularization}, with \textbf{group lasso} for group sparsity and with \textbf{sparse group} regularization (a combination of $\ell_1$ regularization and group lasso -- see Eq. \ref{eqn:final-obj}) for both overall sparsity and group sparsity. As shown in Table \ref{tab:results_sparsity}, the sparse group regularizer outperforms.

We also explore feature selection functionality by investigating the magnitudes of elements in matrix $\bm{\Theta}$. The element $\Theta_{ij}$ measures the influence of profile feature or marker $j$ to label $i$. Small (large) values indicate the corresponding features or markers have little (high) influence to label. For example, in \emph{LinkedIn-Career} dataset, the numerical values in coefficient column vector corresponding to marker $engineer$ are all nonzero, showing that having a working experience as an $engineer$ is important in IT industry. For marker $director$, most of the elements in the corresponding coefficient column vector is zero except for the rows of positions $CEO$ and $founder$, suggesting an ascending career path in general.

\section{Conclusion}
We study the problem of factorial point process learning for which the event can carry multiple markers whereby the relevant concept can be found in  Factorial Hidden Markov Models \cite{ghahramani1996factorial}. Two learning algorithms are presented: the first is directly based on the raw regularized discriminative prediction objective function which employs ADMM and FISTA techniques for optimization; the second is a simple LR solver which is based on a key reformulation of the raw objective function. Experimental results on two real-world datasets collaborate the effectiveness of our approach.
\section*{Acknowledgments}
The work is partially supported by National Natural Science Foundation of China (61602176, 61628203, 61672231), The National Key Research and Development Program of China (2016YFB1001003), NSFC-Zhejiang Joint Fund for the Integration of Industrialization and Informatization (U1609220).
\bibliographystyle{IEEEtran}
\bibliography{IEEEfull}
\vspace{-1cm}
\begin{IEEEbiography}[{\includegraphics[width=1in,height=1.25in,clip,keepaspectratio]{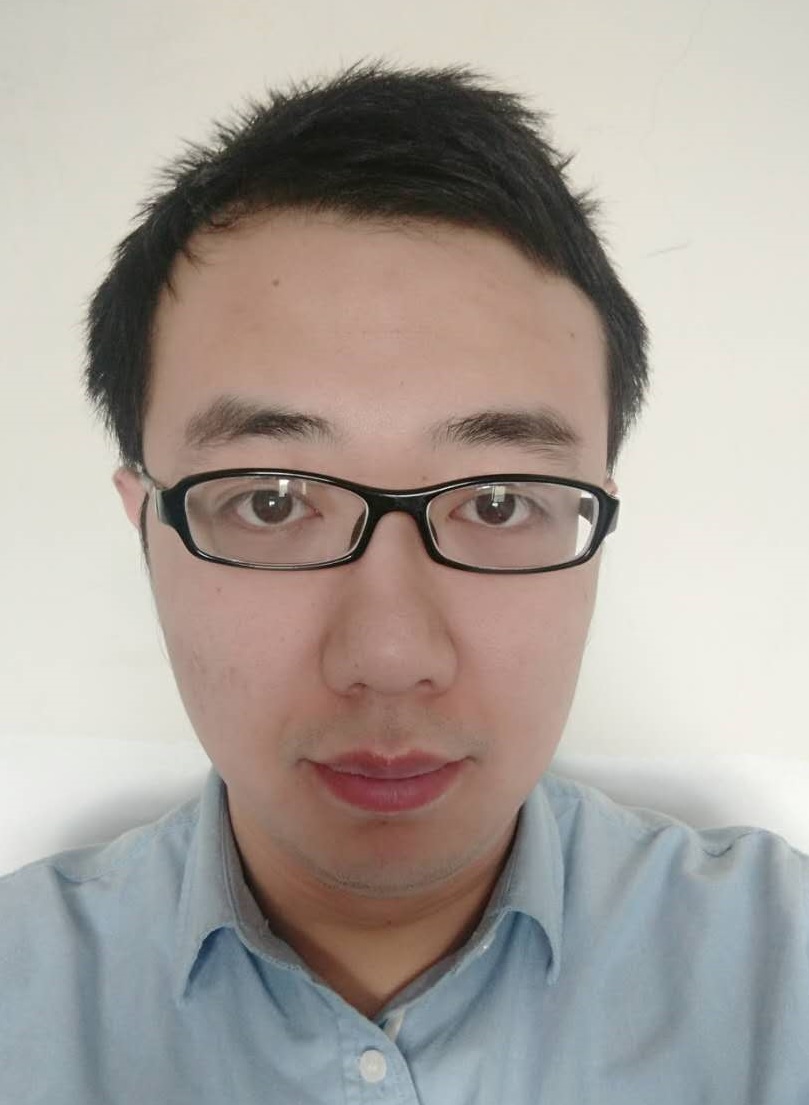}}]{Weichang Wu}
	received the B.S. degree in electronic engineering from Huazhong University of Science and Technology, Wu Han, China, in 2013. He is currently pursuing the Ph.D. degree with the Department of Electronic Engineering, Shanghai Jiao Tong University, Shanghai, China. His current research interests include data mining, especially medical information mining and disease modeling based on event sequence learning.
\end{IEEEbiography}
\vspace{-1cm}
\begin{IEEEbiography}[{\includegraphics[width=1in,height=1.25in,clip,keepaspectratio]{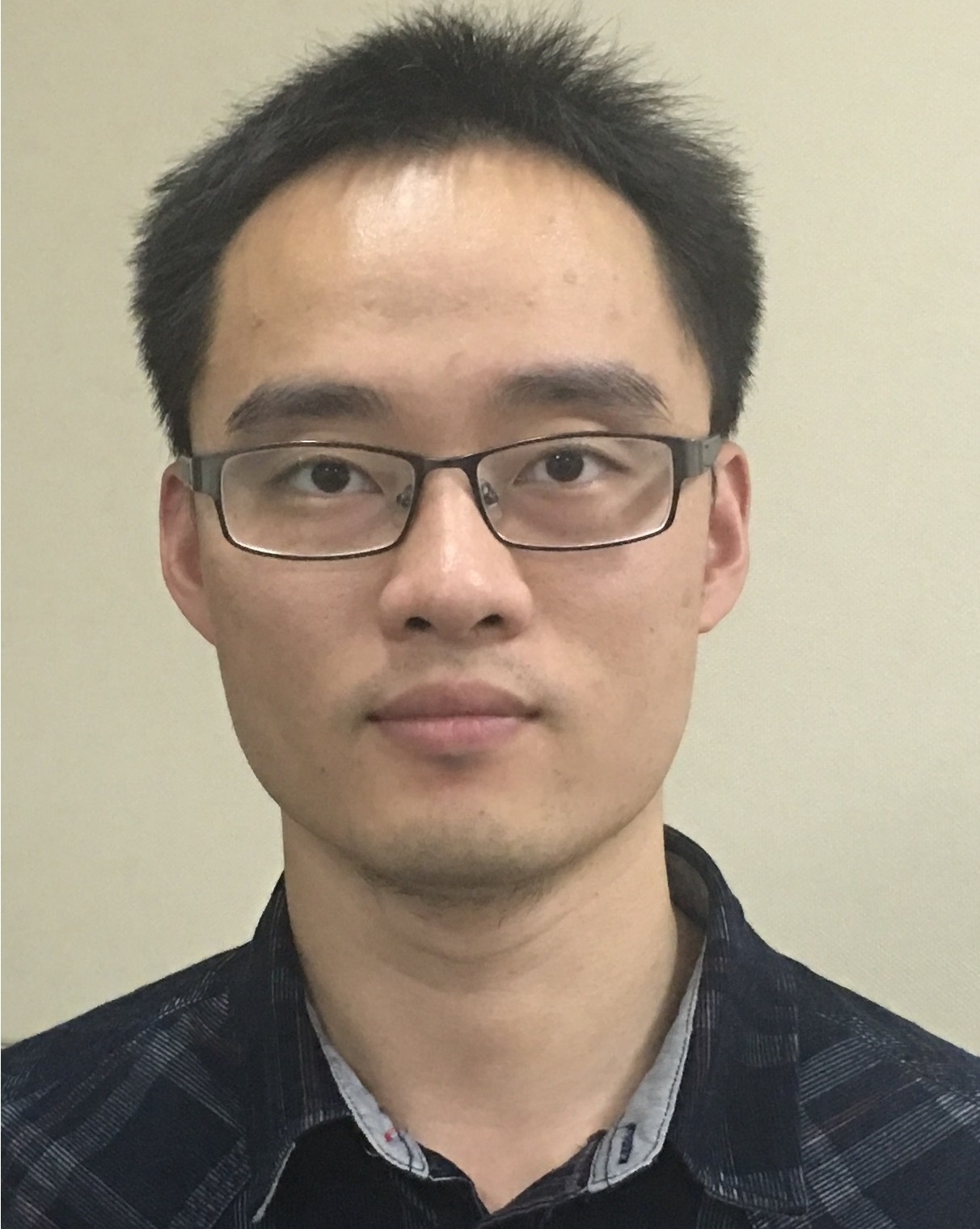}}]{Junchi Yan}
	(M'10) is currently an Associate Professor with Shanghai Jiao Tong University. Before that, he was a Senior Research Staff Member and Principal Scientist for visual computing with IBM Research where he started his career since April 2011. He obtained the Ph.D. at the Department of Electronic Engineering of Shanghai Jiao Tong University, China. He received the ACM China Doctoral Dissertation Nomination Award and China Computer Federation Doctoral Dissertation Award. His research interests are machine learning and visual computing. He serves as an Associate Editor for IEEE ACCESS and on the executive board of ACM China Multimedia Chapter.
\end{IEEEbiography}
\vspace{-1cm}
\begin{IEEEbiography}[{\includegraphics[width=1in,height=1.25in,clip,keepaspectratio]{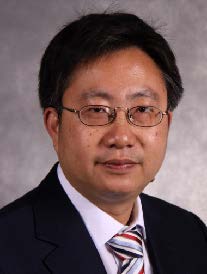}}]{Xiaokang Yang} (M'00-SM'04) received the B. S.
	degree from Xiamen University, Xiamen, China, in 1994, the M. S. degree from Chinese Academy of Sciences in 1997, and the Ph.D. degree from Shanghai Jiao Tong University in 2000. He is currently a Distinguished Professor of School of Electronic Information and Electrical Engineering, Shanghai
	Jiao Tong University, Shanghai, China. His research interests include visual signal processing and communication, media analysis and retrieval, and pattern recognition. He serves as an Associate Editor of IEEE Transactions on Multimedia and an Associate Editor of IEEE Signal	Processing Letters.
\end{IEEEbiography}
\vspace{-1cm}
\begin{IEEEbiography}[{\includegraphics[width=1in,height=1.25in,clip,keepaspectratio]{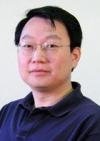}}]{Hongyuan Zha} is a Professor at the School of
	Computational Science and Engineering, College of Computing, Georgia Institute of Technology and East China Normal University. He earned his PhD
	degree in scientific computing from Stanford University in 1993. Since then he has been working on information retrieval, machine learning applications and numerical methods. He is the recipient of the Leslie Fox Prize (1991, second prize) of the Institute of Mathematics and its Applications, the Outstanding Paper Awards of the 26th International Conference on Advances in Neural Information Processing Systems (NIPS
	2013) and the Best Student Paper Award (advisor) of the 34th ACM SIGIR International Conference on Information Retrieval (SIGIR 2011). He was an
	Associate Editor of IEEE Transactions on Knowledge and Data Engineering.
\end{IEEEbiography}
\end{document}